%% file: main.tex
\documentclass[10pt,twocolumn,letterpaper]{article}

\usepackage{iccv}              
\usepackage[margin=1in]{geometry}
\usepackage{multirow}
\usepackage{makecell}
\usepackage{booktabs}
\usepackage{colortbl}
\usepackage{xcolor}
\usepackage{graphicx}


\definecolor{iccvblue}{rgb}{0.21,0.49,0.74}
\usepackage[pagebackref,breaklinks,colorlinks]{hyperref}
\hypersetup{
    citecolor=green,     
    linkcolor=blue,       
    urlcolor=iccvblue    
}

\input{src/others/env.tex}

\def\paperID{10238} 
\def\confName{ICCV}
\def\confYear{2025}

\title{GMAI-VL-R1: Harnessing Reinforcement Learning \\ for Multimodal Medical Reasoning}

\author{
Yanzhou Su\footnotemark[1]\ , Tianbin Li\thanks{Equal contribution}\ , Jiyao Liu, Chenglong Ma, Junzhi Ning, Cheng Tang \\
Sibo Ju, Jin Ye, Pengcheng Chen, Ming Hu, Shixiang Tang, Lihao Liu, Bin Fu \\
Wenqi Shao, 
Xiaowei Hu,
Xiangwen Liao\footnotemark[2], 
Yuanfeng Ji\footnotemark[2], 
Junjun He\thanks{Corresponding author} \\   
Fuzhou University \quad Shanghai Artificial Intelligence Laboratory \quad Shanghai Innovation Institute \\ 
Fudan University \quad Monash University \quad University of Washington \quad Stanford University
}

\begin{document}
\maketitle
\input{src/secs/0_abstract}
\input{src/secs/1_intro}

\input{src/secs/2_related_work}

\input{src/secs/3_methodology}

\input{src/secs/4_experiments}

\input{src/secs/5_conclusion}

\newpage
{
    \small
    \bibliographystyle{ieeenat_fullname}
    \bibliography{src/others/library.bib}
}

\clearpage
\appendix
\input{src/secs/8_appendix}

\end{document}

%% file: src/others/env.tex
\usepackage{algorithm}
\usepackage{algpseudocode}

\usepackage[capitalize,noabbrev]{cleveref}
\usepackage{appendix}
\usepackage{cleveref}
\usepackage{xcolor}
\usepackage{pifont}
\crefname{section}{Section}{Sections}
\crefname{theorem}{Theorem}{Theorems}
\crefname{lemma}{Lemma}{Lemmas}
\crefname{equation}{Equation}{Equations}
\crefname{proposition}{Proposition}{Propositions}
\crefname{claim}{Claim}{Claims}
\crefname{appendix}{Appendix}{Appendices}
\crefname{algorithm}{Algorithm}{Algorithms}
\crefname{figure}{Figure}{Figs}
\crefname{table}{Table}{Tables}
\crefname{remark}{Remark}{Remarks}
\crefname{definition}{Definition}{Definitions}
\crefname{equation}{Equation}{Equations}
\crefname{corollary}{Corollary}{Corollaries}
\crefalias{section}{appendix}

\definecolor{codegreen}{rgb}{0,0.6,0}
\definecolor{codegray}{rgb}{0.5,0.5,0.5}
\definecolor{codepurple}{rgb}{0.58,0,0.82}
\definecolor{backcolour}{rgb}{0.95,0.95,0.92}
\definecolor{lightgreen}{HTML}{30E1C8}
\definecolor{lightblue}{HTML}{0254D6}
\definecolor{cite_color}{HTML}{114083}
\definecolor{link_color}{RGB}{153, 0,0}  
\definecolor{url_color}{RGB}{153, 102,  0}
\definecolor{emp_color}{RGB}{0,0,255}
\definecolor{myyellow}{HTML}{FFFADF}

\newcommand{\tablestyle}[2]{
    \setlength{\tabcolsep}{#1}
    \renewcommand{\arraystretch}{#2}
}

\newcommand{\xmark}{\ding{55}}

%% file: src/secs/0_abstract.tex
\begin{abstract}
Recent advances in general medical AI have made significant strides, but existing models often lack the reasoning capabilities needed for complex medical decision-making. 
This paper presents GMAI-VL-R1, a multimodal medical reasoning model enhanced by reinforcement learning (RL) to improve its reasoning abilities. 
Through iterative training, GMAI-VL-R1 optimizes decision-making, significantly boosting diagnostic accuracy and clinical support. 
We also develop a reasoning data synthesis method, generating step-by-step reasoning data via rejection sampling, which further enhances the model’s generalization. 
Experimental results show that after RL training, GMAI-VL-R1 excels in tasks such as medical image diagnosis and visual question answering. 
While the model demonstrates basic memorization with supervised fine-tuning, RL is crucial for true generalization. 
Our work establishes new evaluation benchmarks and paves the way for future advancements in medical reasoning models. 
Code, data, and model will be released at \href{https://github.com/uni-medical/GMAI-VL-R1}{this link}.

\end{abstract}

%% file: src/secs/1_intro.tex
\section{Introduction}
\label{sec:intro}


\begin{figure}
    \centering
    \includegraphics[width=\linewidth]{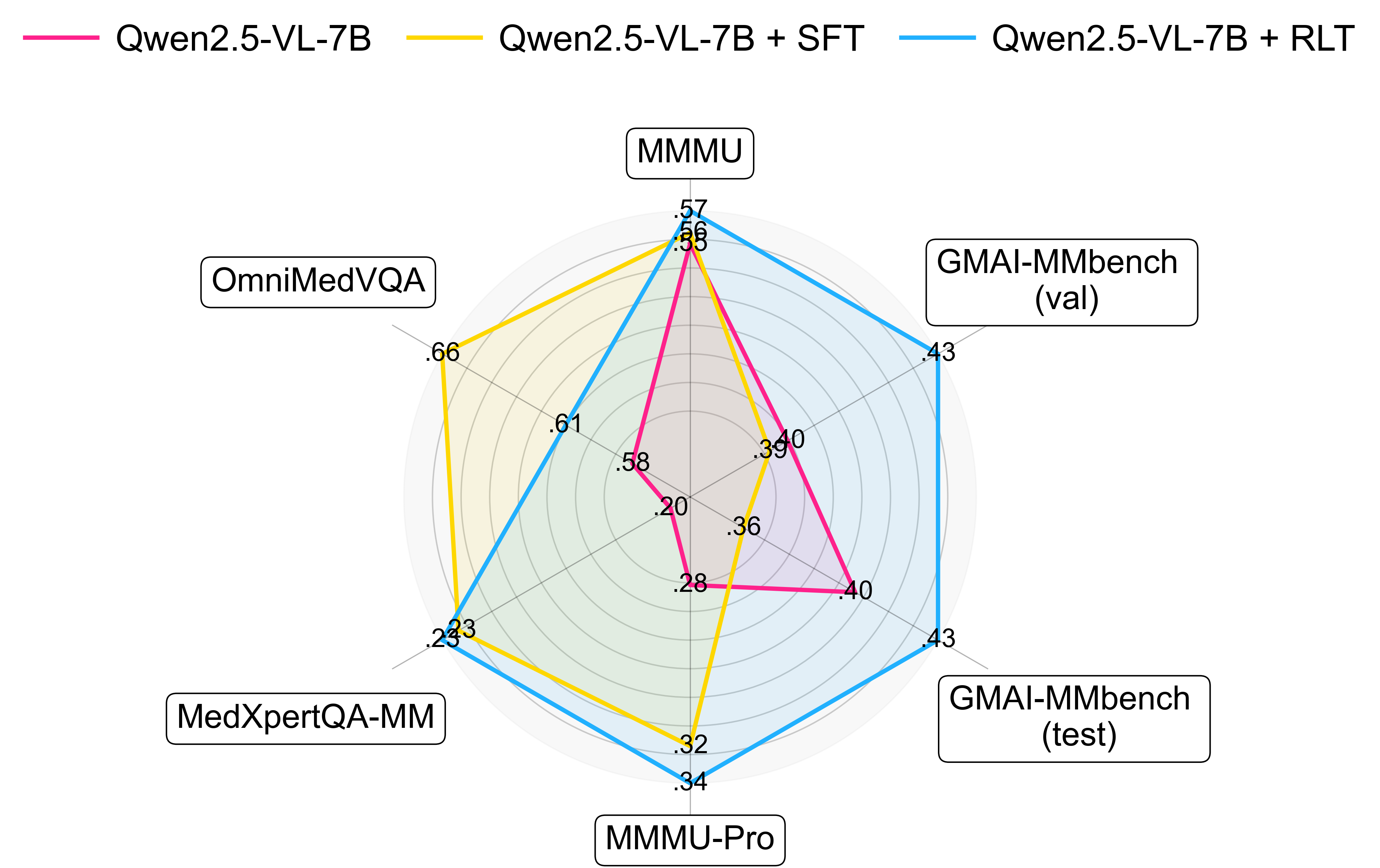}
    \caption{\textbf{Quantitative comparison of the performance of different models across various benchmarks}. The results show that, in most benchmarks, the RLT-based model outperforms the SFT-based model.}
    \label{fig:model-comparison}
\end{figure}

Integrating multimodal medical data such as images, clinical records, and patient histories is crucial for improving healthcare quality and efficiency.
Multimodal models leverage this diverse medical information to support comprehensive decision making, enhance diagnostic accuracy, and improve clinical outcomes~\cite{acosta2022multimodal,lipkova2022artificial} 
These models are especially valuable in complex clinical settings where information from multiple sources must be processed simultaneously.
Nevertheless, significant challenges persist in developing large-scale multimodal models capable of effectively reasoning about medical data, particularly when advanced reasoning and reflective capabilities are essential for precise clinical decisions~\cite{kline2022multimodal}.
These challenges have encouraged the creation of increasingly sophisticated models to provide better diagnostic support.

Existing medical multimodal models have significantly advanced through fine tuning on carefully constructed multimodal instruction data~\cite{li2024gmai, chen2024huatuogpt, moor2023med, tu2024towards, li2024llava, liu2023qilin, zhang2024generalist}.
These instructionally tuned models excel at diagnostic tasks, enabling quicker and more accurate disease detection, for instance, diabetic retinopathy screening~\cite{li2024integrated} and pneumonia detection~\cite{thawakar2024xraygpt}.
By integrating medical image and text data pairs, they outperform single modality models across tasks such as image captioning, visual question answering, and medical report generation.
However, a key limitation is their limited reasoning capability.
Most existing models depend heavily on supervised fine tuning (SFT), emphasizing memorization of input and output mappings~\cite{chu2025sft}, rather than developing deeper reasoning abilities.
Although these models perform well on familiar tasks, they lack sufficient flexibility when encountering novel or complex scenarios.
In medical contexts, where data complexity and uncertainty are prevalent, relying only on pattern recognition is inadequate for effective clinical decision making.

\begin{figure}[t]
    \centering
    \includegraphics[width=\linewidth]{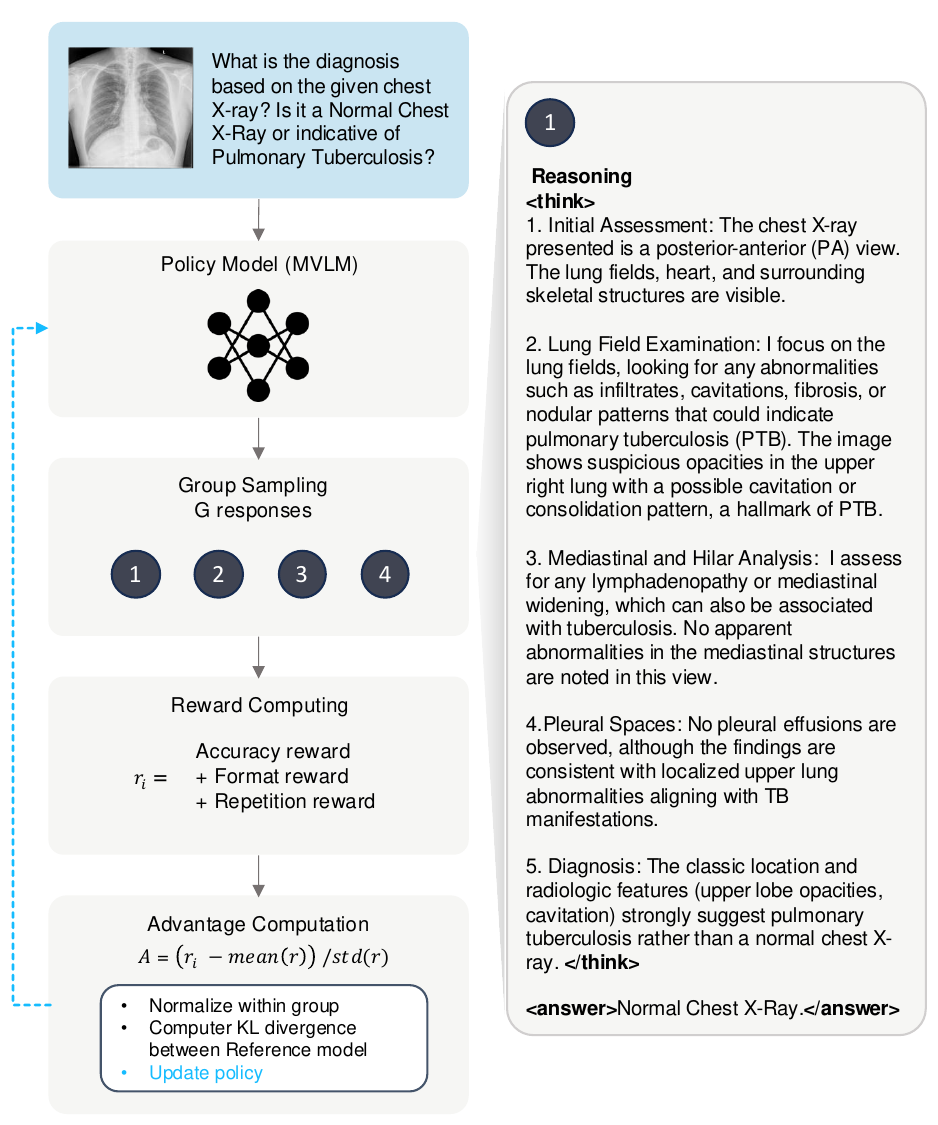}
    \caption{\textbf{The framework of reinforcement learning tuning}. Given the input medical image (chest X-ray) and question, the policy model generates multiple reasoning responses through group sampling. The reasoning steps are then evaluated based on accuracy, format, and repetition rewards.  This process updates the policy model, guiding it towards more accurate diagnoses. The final answer, ``Normal Chest X-ray,'' is generated based on the reasoning process that identifies pulmonary tuberculosis indicators.}
    \label{fig:rl_pipeline}
    \vspace{-5mm}
\end{figure}

Inspired by DeepSeek-R1~\cite{deepseekr1}, we develop GMAI-VL-R1, a general purpose multimodal medical model leveraging reinforcement learning tuning (RLT) to enhance Chain of Thought (CoT) reasoning and reflection~\cite{wei2022chain}.
Unlike traditional models that depend on supervised fine tuning (SFT), GMAI-VL-R1 directly applies RLT to the base model, specifically Qwen VL 7b~\cite{bai2023qwen}. The RLT training procedure is illustrated in Fig.~\ref{fig:rl_pipeline}.
Given a medical image, GMAI-VL-R1 generates multiple responses, each including explicit reasoning steps and final answers produced by large vision language models (LVLMs). We then utilize rule based rewards, including accuracy, format correctness, and redundancy avoidance, to guide policy gradient optimization~\cite{shao2024grpo} during model updates.
Through reinforcement learning tuning, the model gains practical reasoning experience via repeated training and self correction, surpassing superficial pattern recognition.
This strengthened reasoning capability enables GMAI-VL-R1 to provide more reliable results in high risk clinical decisions and equips it to manage complex and previously unseen medical scenarios effectively, as demonstrated in Fig.~\ref{fig:model-comparison}.

To develop our RLT approach, we first carefully constructed GMAI-Reasoning10K, a high-quality medical visual question answering (VQA) dataset (Fig.~\ref{fig:modality_distribution}). 
This dataset contains 10,000 rigorously curated medical VQA pairs derived from 95 public datasets, spanning 12 imaging modalities, such as X-ray, CT, and MRI. 
Unlike contemporary works~\cite{lai2025med} that rely solely on questions and answers as instructions, we generated detailed CoT reasoning instructions for each QA pair to ensure a fair comparison between RLT and SFT approaches.
Specifically, GPT-4o was employed to produce comprehensive reasoning chains, followed by a specialized filtering strategy designed to refine and retain only high-quality reasoning steps. 
This detailed Chain-of-Thought instruction curation ensures that SFT models reach their true upper performance limit, enabling a fair comparison of the genuine reasoning capabilities between RL and SFT approaches.

We conducted comprehensive evaluations across 6 large-scale medical multimodal benchmarks~\citep{hu2024omnimedvqa,chen2024gmai,yue2024mmmu,wang2024mmlu,medxpert} to rigorously assess GMAI-VL-R1's performance.
Results from multiple medical benchmarks demonstrate that GMAI-VL-R1 consistently excels in medical question answering, disease diagnosis, and recognition tasks, highlighting reinforcement learning's crucial role in enhancing medical perception and reasoning capabilities.
Notably, GMAI-VL-R1 outperforms traditional SFT methods in generalization, emphasizing its significant potential for real-world clinical applications.
Fig.~\ref{fig:model-comparison} illustrates this advantage clearly, showing the substantial improvements achieved by our RLT-enhanced model in both familiar and previously unseen medical scenarios.

Specifically, on the MMMU benchmark, GMAI-VL-R1 achieves a accuracy of 57.33\%, representing a 2\% improvement over the baseline model. 
For MMMU-pro, the model attains an accuracy of 34.03\%, yielding a substantial gain of 5.56\%. 
On GMAL-MMbench, GMAI-VL-R1 reaches accuracies of 43.14\% on the validation set and 43.84\% on the test set, corresponding to improvements of 3.12\% and 3.25\% over the baseline, and in both cases, it remains 3.48\% and 7.72\% higher than the SFT model.  
Furthermore, the method obtains 23.80\%  (+3.50\%) on MedXpertQA-MM and 61.01\% (+2.60\%) on OmniMedVQA. 
These empirical findings conclusively demonstrate that our proposed RLT strategy consistently outperforms both the baseline model and the SFT method across various medical multimodal tasks, highlighting its potential as a versatile optimization technique for medical vision-language models.

Overall, our contributions in this work are three-fold:

\begin{itemize}
    \item We develop GMAI-VL-R1, a multimodal medical model that directly applies reinforcement learning tuning to enhance reasoning capabilities, moving beyond pattern memorization in medical AI. 
    \item We construct GMAI-Reasoning10K, a high-quality dataset spanning 12 imaging modalities with detailed Chain-of-Thought annotations, establishing a reliable benchmark for developing and comparing tuning approaches.
    \item We demonstrate through extensive experiments across multiple benchmarks that our approach achieves competitive performance compared to previous state-of-the-art methods, achieves superior generalization in out-of-distribution scenarios, while requiring significantly less training data.
\end{itemize}

%% file: src/secs/2_related_work.tex
\section{Related Work}
\paragraph{Medical Vision-Language Models} are typically built on general-purpose large models and adapted to medical applications using specialized datasets~\cite{Hartsock2024}. For example, Med-Flamingo~\citep{moor2023med} enhances OpenFlamingo-9B with 800K medical image-text pairs, focusing on medical image analysis and report generation. RadFM~\citep{wu2023towards} improves PMC-LLaMA~\citep{wu2023pmc} with 16 million radiology images and text. Med-PaLM~\citep{tu2024towards} adapts PaLM-E~\citep{driess2023palm} for medical tasks using one million samples, excelling in diagnostics and Q\&A. LLaVA-Med~\citep{li2024llava} uses PubMed Central data to enhance LLaVA~\citep{touvron2023llama}, improving biomedical image understanding and open-domain conversation. MedTrinity-25M~\citep{xie2024medtrinity} creates 25M image-text pairs for fine-tuning, though results remain limited. Qilin-Med-VL~\cite{liu2023qilin} and BioMedGPT~\cite{zhang2024generalist} also use large image-text pairs, but their performance is limited by data quality, hindering generalization. Med-Gemini~\citep{saab2024capabilities} enhances Gemini with long-format Q\&A datasets, improving performance in complex medical reasoning tasks.

Despite significant progress, existing medical multimodal models still face limitations in both data and training methods. For instance, these models typically rely on limited medical datasets, which hinder their ability to generalize effectively, especially in complex medical scenarios that require reasoning and decision-making. To address these challenges, we leverage high-quality medical reasoning data and apply reinforcement learning (RL) to enhance the reasoning and decision-making capabilities of medical multimodal models.

\paragraph{Reasoning Models}
Reasoning models are essential in multimodal learning, particularly for tasks that require integrating modalities like images and text for decision-making. As multimodal reasoning demand grows, frameworks and techniques have evolved. The introduction of OpenAI-O1~\cite{jaech2024openai} marked a key development, though it still struggles with complex vision-language tasks. Chain-of-Thought (CoT)\cite{wei2022chain} breaks reasoning into steps, improving capability, while Progressive Reasoning Models (PRM)\cite{lightman2023let} optimize outcomes through intermediate supervision. Sky-T1~\cite{sky_t1_2025} enhances high-level reasoning with supervised fine-tuning (SFT), and LIMA~\cite{zhou2023lima} proves that limited SFT data can still achieve strong reasoning performance. Similarly, \cite{huang2025o1} shows minimal SFT data can boost reasoning.


However, while SFT primarily optimizes the structure and logic of reasoning, its capacity for more complex tasks remains limited. In response, recent advancements in reinforcement learning (RL)-based reasoning models have shown significant progress~\cite{ziegler2019finetuning, ouyang2022training, zhou2024archer, deepseekr1}. For example, DeepSeek-R1~\cite{deepseekr1} leverages RL to refine the reasoning process, improving efficiency and accuracy in complex tasks. The Qwen-QwQ~\cite{qwq-32b-preview} model expands reasoning capabilities, particularly in long-context reasoning, by utilizing large-scale context processing and RL to deepen the model’s reasoning abilities.
Moreover, Kimi-R1.5~\cite{team2025kimi} introduced an innovative framework combining long-chain-of-thought (long-CoT) with RL, significantly enhancing performance in multimodal reasoning tasks and demonstrating potential for more complex tasks.

Thus, the combination of SFT and RL not only strengthens the fundamental logic and structure of reasoning but also enhances knowledge generalization and flexibility through RL's strategy adjustments~\cite{chu2025sft}. This combined approach allows the model to perform effective reasoning and decision-making with limited data, expanding its capability in more complex tasks.

%% file: src/secs/3_methodology.tex
\section{Methodology} 
\label{sec:method}

To explore reinforcement learning tuning in medical multimodal analysis, we first curate a high-quality visual question-answering dataset tailored for medical image analysis. 
This dataset was developed using an automatic pipeline that ensures both data quality and diversity.
We selected Qwen-VL as our base model because of its strong language generation capabilities and proven potential for \textit{self-improvement}.
We implemented tuning across different model sizes (3B and 7B) and conducted experiments on various benchmarks.
Our reinforcement learning tuning framework builds upon the Group Relative Policy Optimization (GRPO) method~\cite{shao2024grpo}, adopting a similarly straightforward design.

\begin{figure}[t]
    \centering
    \includegraphics[width=\linewidth]{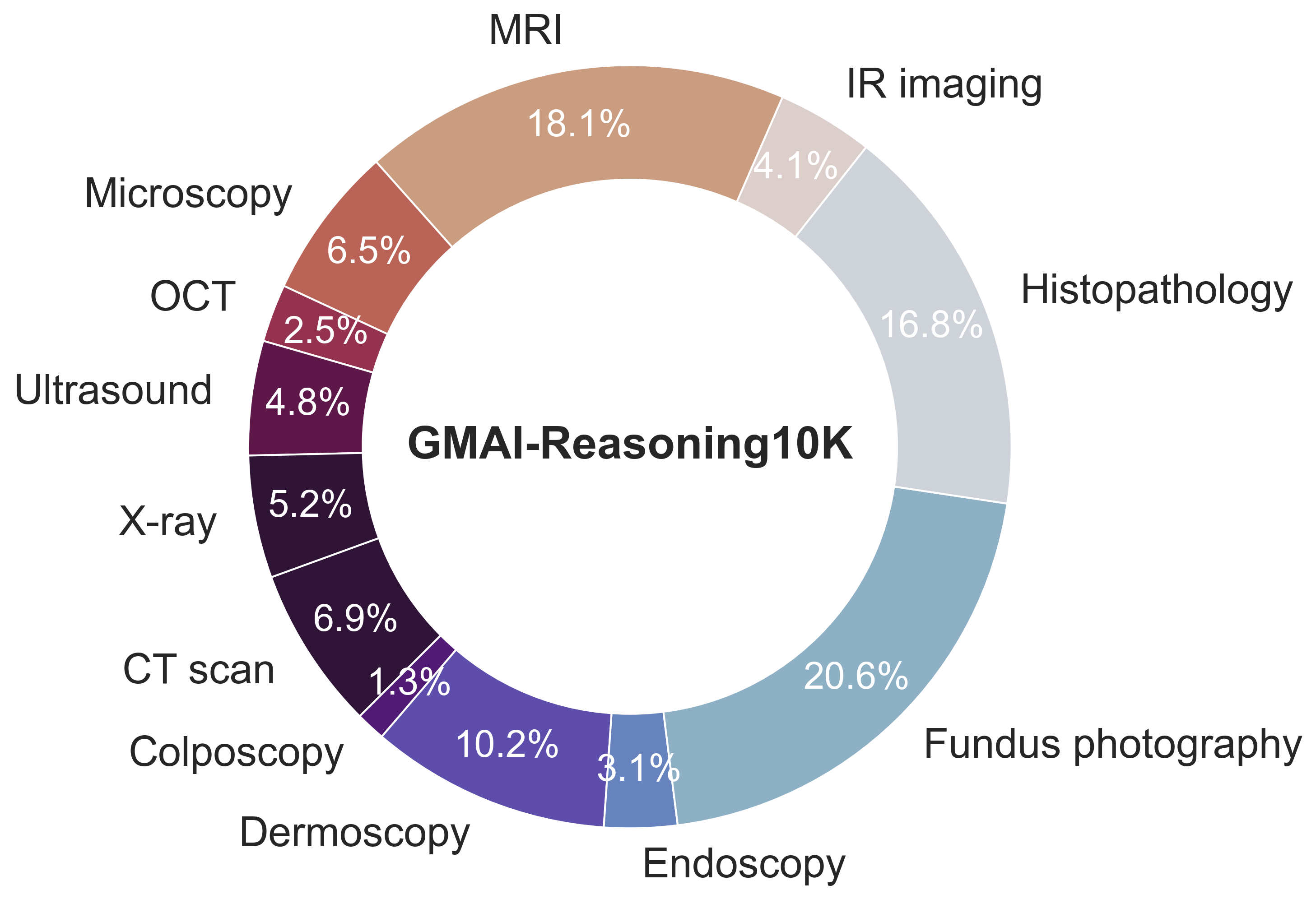}
    \caption{\textbf{Modality distribution of the curated GMAI-Reasoning10K dataset.} GMAI-Reasoning10K provides high-quality 10K visual question answering pairs spanning 12 different medical modalities.}
    \label{fig:modality_distribution}
    \vspace{-5mm}
\end{figure}

\subsection{GMAI-Reasoning10K}

\begin{figure*}[h]
    \centering
    \includegraphics[width=\linewidth]{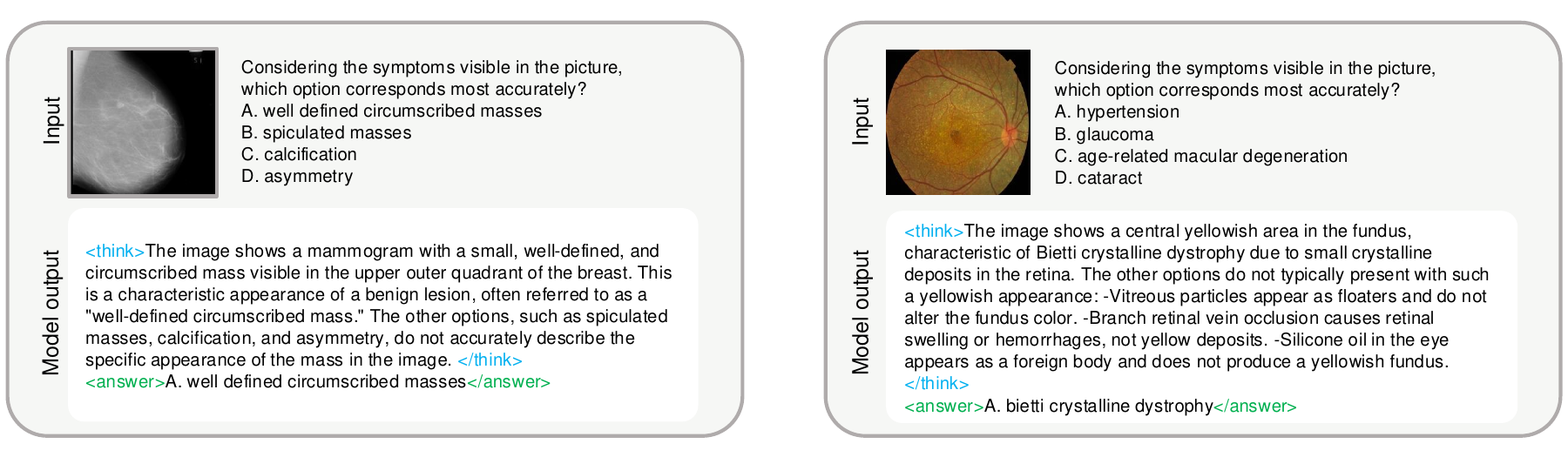}
    \caption{\textbf{Case study illustrating the model's reasoning ability under Reinforcement Learning Tuning (RLT)}. Given medical images, the model identifies the most accurate diagnosis based on visible symptoms. RLT encourages the model to engage in reasoning and select the correct answer from multiple choices.}
    \label{fig:example_output}
\end{figure*}

\paragraph{Visual Question Answering.}
Our dataset construction process began with the collection of multimodal medical data from 95 datasets from reliable sources such as Kaggle, GrandChallenge, and Open-Release, covering 12 imaging modalities (e.g., X-ray, CT, MRI). 
The data preprocessing pipeline was based on methods from SAMed-20M~\cite{ye2023sa}, which included augmentation and standardization; for instance, 3D data (CT/MRI) had individual slices extracted and pixel values normalized to 0–255, while video data was processed by extracting frames at 2 frames per second. 
Key metadata, such as background information, modality, and labels, was extracted from each dataset and subsequently used to construct informative prompts via GPT for generating multiple-choice questions (with a single correct answer). 
To ensure quality, a reject sampling strategy was applied to discard samples that did not meet our predefined criteria, such as relevant annotations, or correct labels, resulting in a high-quality dataset comprising 10K samples. 
This data distillation approach has also been adopted by numerous studies~\cite{deepseekr1,sky_t1_2025,liu2025visual}.
It should be noted that our data selection avoids any overlap with the test data used in commonly-used benchmarks.
The distribution of this dataset is illustrated in \cref{fig:modality_distribution}, and more curation details can be found in the supplementary material.

\paragraph{Chain-of-Thought Instruction.}

In contrast to concurrent RFT approaches~\cite{liu2025visual} in general vision tasks that directly use questions and answers as instructions for SFT, we constructed detailed chain-of-thought (COT) explanations for each VQA pair to ensure a fair comparison between RFT and SFT.
Specifically, for each VQA pair, we prompted GPT-4o to generate a comprehensive COT for each VQA pair, see the Appendix for detailed prompt information. 
Similarly, we used a special curation strategy to filter and refine high-quality COTs.
This process ultimately resulted in the creation of the GMAI-Reasoning10K dataset, which comprises 10K VQA pairs along with their corresponding reasoning COTs, serving as a resource for developing and comparing instruction-tuning-based methods.

\begin{algorithm}[t]
    \footnotesize
    \caption{Reinforcement Learning Tuning}
    \label{tab:grpo}
    \begin{algorithmic}[1]
    \State \textbf{Input:} Medical image \(I\), question \(q\), and answer \(x\)
    \State Define policy model \( \pi_{\theta}^{\text{RL}} \) and reference model \( \pi_{\text{ref}} \)
    \State Define reward: \( r(x,q,y) = r_{\text{acc}} + r_{\text{fmt}} + r_{\text{rep}} \)
    \For{each iteration \(t\)}
        \For{each input \((x_i, q_i)\)}
            \State Sample outputs: \( \{ y_i^j \}_{j=1}^{k} \sim \pi_{\theta_{\text{RL}}^t}(x_i, q_i) \)
            \State Compute advantage:
            {
            \setlength{\abovedisplayskip}{-0.32em}%
            \setlength{\belowdisplayskip}{-0.32em}%
            \[
            A(x_i, y_i^j) = r(x_i, y_i^j) - \frac{1}{k}\sum_{l=1}^{k} r(x_i, y_i^l) 
            \]
            }
        \EndFor
        \State GRPO loss: 
        {
        \setlength{\abovedisplayskip}{0.15em}%
        \setlength{\belowdisplayskip}{0.15em}%
        \[
        \mathcal{L}_{\text{GRPO}} = -\mathbb{E}\Bigl[A(x,q,y) \cdot \min\Bigl(\frac{\pi_{\theta}(y|x,q)}{\pi_{\theta_{\text{RL}}^t}(y|x,q)}, 1+\epsilon\Bigr)\Bigr]
        \]}
        \State Total loss: \( \mathcal{L}_{\text{RL}} = \mathcal{L}_{\text{GRPO}} + \beta \cdot D_{KL}\bigl(\pi_{\theta} \parallel \pi_{\text{ref}}\bigr) \)
        \State Update: \( \theta_{\text{RL}}^{t+1} \leftarrow \theta_{\text{RL}}^t - \alpha \nabla_{\theta} \mathcal{L}_{\text{RL}} \)
    \EndFor
    \State \Return Final model \( \pi_{\theta_{\text{RL}}} \)
    \end{algorithmic}
\end{algorithm}

\subsection{Reinforcement Learning Tuning.}
Our Reinforcement Learning Tuning (RLT) pipeline is shown in \cref{fig:rl_pipeline}, which is based on Group Relative Policy Optimization (GRPO) framework.
We are the first to apply GRPO to the multimodal medical domain, with extensive validation conducted at scale.
We apply this simple yet effective framework to the base model (e.g., Qwen2.5-VL-7b~\cite{Qwen2.5-VL}), which enables the model to develop advanced medical reasoning capabilities on its own without relying on supervised data.

\paragraph{Preliminaries.}

Let \( I \) denote a medical image and \( (q, x) \) the associated VQA pair. We initialize a reference model \( \pi_{\text{ref}} \) and a reinforcement learning model \( \pi_{\theta}^{\text{RL}} \) (both obtained from base model), which will be optimized via our RLT pipeline.
%
The RLT pipeline begins with the base Q\&A pair \( (q, x) \), where \( q \) represents the question and \( x \) denotes the ground truth answer. The baseline policy, \( \pi_{\text{ref}} \), guides the learning of the reinforcement learning policy, \( \pi_{\theta}^{\text{RL}} \), through a carefully designed reward mechanism.

\paragraph{Group Sampling.}  
During each iteration \( t \), for each input \( q_i \), we sample a group of \( k \) outputs \( \{ y_i^j \}_{j=1}^{k} \) from the current policy \( \pi_{\theta^t}^{\text{RL}}(q_i) \). 
Group sampling enables the model to explore diverse responses for the same input, which is essential for robust policy evaluation.

\paragraph{Reward Function.}  
Each output is evaluated using a reward function \( r(x, y) \) that encapsulates three key components:
$r(x, y) = r_{\text{acc}}(x, y) + r_{\text{fmt}}(x, y) + r_{\text{rep}}(x, y)$,
where: 
\begin{itemize}
    \item Accuracy Reward (\( r_{\text{acc}} \)): Assesses whether the response is correct (e.g., matching a standard answer for multiple-choice questions). 
    \item Format Reward (\( r_{\text{fmt}} \)): Ensures that the model encloses its reasoning within designated tags (e.g., \texttt{<think>} and \texttt{</think>}) and wraps the final answer within \texttt{<answer>} and \texttt{</answer>} tags.
    \item Repetition Penalty Reward (\( r_{\text{rep}} \)): Penalizes repetitive or redundant outputs to encourage concise and diverse reasoning.
\end{itemize}




\paragraph{Advantage Computation.}  
For each sampled output \( y_i^j \), we compute the advantage function:
\begin{equation}
A(x_i, y_i^j) = r(x_i, y_i^j) - \frac{1}{k}\sum_{l=1}^{k} r(x_i, y_i^l),
\end{equation}
which quantifies the relative performance of each output compared to the average reward of the group.

\paragraph{Policy Optimization.}  
The core of our approach is the GRPO loss, defined as:
\begin{equation}
\mathcal{L}_{\text{GRPO}} = -\mathbb{E}\left[ A(x, y) \cdot \min\left( \frac{\pi_{\theta}(y|x)}{\pi_{\theta_{\text{RL}}^t}(y|x)}, 1+\epsilon \right) \right],
\end{equation}
where the clipping operation ensures stability in the updates. To regularize the learning process and prevent drastic deviations from the supervised baseline, we include a KL divergence term:
\begin{equation}
\mathcal{L}_{\text{RL}} = \mathcal{L}_{\text{GRPO}} + \beta \cdot D_{KL}(\pi_{\theta} \parallel \pi_{\text{ref}}).
\end{equation}
The hyperparameter \( \beta \) (initialized to 0.04) acts as a weighting factor to balance the KL divergence regularization term and the main GRPO loss, ensuring a controlled trade-off between the policy model's stability and creativity.
The model parameters are updated using gradient descent:
\begin{equation}
\theta_{\text{RL}}^{t+1} \leftarrow \theta_{\text{RL}}^t - \alpha \nabla_{\theta} \mathcal{L}_{\text{RL}},
\end{equation}
This iterative process continues until convergence, thereby progressively self-improving the model’s medical reasoning capabilities.

\input{src/tables/benchmark_summary}

%% file: src/tables/benchmark_summary.tex
\begin{table}[t]
    \centering
    \setlength{\tabcolsep}{3.5pt}
    \resizebox{\linewidth}{!}{
    \begin{tabular}{l|c|c}
    \hline
    Benchmark            & \# Samples & Out-of-distribution? \\
    \hline
    MMMU(H\&M) \cite{yue2024mmmu}                 & 150                  & \checkmark                  \\
    MMMU-pro \cite{wang2024mmlu}                       & 288
    & \checkmark                 \\
    GMAI-MMBench (\textit{val}) \cite{chen2024gmai}               & 4,550                & \checkmark                      \\
    GMAI-MMBench (\textit{test}) \cite{chen2024gmai}           & 21,281                & \checkmark                      \\
    MedXpertQA-MM \cite{medxpert}               & 2,000                & \checkmark                 \\ \hline
    OmniMedVQA \cite{hu2024omnimedvqa}                & 127,995                & \xmark                 \\ \hline
    \end{tabular}}
    \caption{\textbf{Evaluation benchmarks used in our experiments}. Note that OmniMedVQA and GMAI-Reasoning10K share the same curated dataset distribution, despite being derived from non-overlapping test and training splits.}
    \label{tab:dataset_metrics}
\end{table}

%% file: src/secs/4_experiments.tex
\section{Experiments}

\subsection{Setup}
\input{src/tables/overall}
\paragraph{Benchmarks.} 
We utilized various widely used benchmarks to conduct a large-scale and comprehensive evaluation of RTL's effectiveness in multimodal medical image analysis. 
Our evaluation benchmarks include OmniMedVQA~\citep{hu2024omnimedvqa}, GMAI-MMBench~\citep{chen2024gmai}, MMMU (H\&M)~\citep{yue2024mmmu}, MMMU-pro ~\citep{wang2024mmlu}, and MedXpertQA~\citep{medxpert}, each designed to assess different facets of medical image understanding and question answering.
As summarized in ~\cref{tab:dataset_metrics}, the data counts and distributions are presented. Notably, OmniMedVQA is treated as an \emph{i.i.d.} test set since it shares the same curated dataset distribution as GMAI-Reasoning10K (albeit derived from non-overlapping test and training splits), while the remaining benchmarks are considered out-of-distribution (OOD). Further details on these benchmarks are provided in Sec.~\ref{sec:benchmark} in the Appendix. 



\paragraph{Setting.}
In our experiments, we compare several methods: (i) the base model; (ii) +SFT, which is fine-tuned using the VQA pairs and corresponding CoT instructions from GMAI-Reasoning10K; and (iii) +RLT, where only the VQA pairs from GMAI-Reasoning10K are used for reinforcement learning tuning.
For all experiments, we default to the Qwen-VL-7B model.
Specifically, we apply LLaMAFactory~\citep{zheng2024llamafactory} for the supervised fine-tuning stage. 
During training, we use AdamW as the optimizer with a learning rate of \(1e^{-5}\) following a cosine decay schedule and a batch size of 32. The SFT training is run for 2 epochs.
For the RLT stage, we utilize the repository\footnote{https://github.com/EvolvingLMMs-Lab/open-r1-multimodal} to perform RL training with GRPO, setting the number of generations per group to 7, and the RL training process runs for one epoch.
After training, we leverage VLMEvalKit~\citep{duan2024vlmevalkit} to evaluate model performance on the benchmarks.
We refer readers to \cref{tab:setting} in the Appendix for more details of the network training.



\subsection{Main Results}

\paragraph{RLT Generalizes Better.}

We first explored supervised fine-tuning SFT and reinforcement learning tuning RLT on the base model. 
As shown in \cref{tab:overall_comparsion}, our ablation study reveals that while SFT serves as a strong baseline for in-distribution tasks, it surpasses the base model by a large margin on OmniMedVQA 66.34\% versus 58.41\%. 
However, it likely capitalizes on spurious correlations or shortcuts to achieve high performance without genuine reasoning. 
In other words, SFT appears to rely on memorizing and directly mapping input-output pairs, which works well when the training and test data share the same distribution. 
However, this approach lacks deeper reasoning capabilities and fails to generalize robustly to out-of-distribution tasks. 
For instance, on benchmarks such as GMAI-MMbench, SFT leads to performance degradation compared to the base model, dropping from 40.02\% to 39.65\% on GMAI-MMbench (\textit{val}) and from 40.59\% to 36.12\% on GMAI-MMbench (\textit{test}).

In contrast, the reinforcement learning tuning approach significantly improves performance across both in-distribution and out-of-distribution tasks. 
Despite the in-distribution benchmarks such as OmniMedVQA, it improves on the base model by 2.60\% percent, reaching 61.01\% percent, although it remains lower than SFT. 
However, on out-of-distribution benchmarks, RLT demonstrates a clear advantage. 
It surpasses both the base and SFT models on GMAI-MMbench val and test by +3.12\% and +3.25\% respectively, and achieves notable gains on MMMU and MMMU-pro with improvements of +2.00\% and +5.56\%. 
Similarly, on MedXpertQA-MM, RLT improves performance by +3.50\% over the base model. These results indicate that reinforcement learning tuning forces the model to actively engage in reasoning and exploration during training, leading to more robust and transferable reasoning abilities across diverse and complex tasks. 
While SFT is effective for tasks that rely on in-distribution pattern recognition, RLT cultivates deeper and more generalizable reasoning capabilities, making it better suited for handling out-of-distribution challenges.
This finding is consistent with some recent observations that RLT generalizes while SFT memorizes~\cite{chu2025sft,liu2025visual}.

\paragraph{RLT Works Efficiently.}
We compared our RLT approach with current state-of-the-art models that employ extensive supervised fine-tuning data. 
For example, while models such as Med-Flamingo (7B) and LLaVA-Med (7B) rely on millions of annotated QA pairs, and larger models like HuatuoVision (34B) and MedDR (40B) benefit from vast amounts of SFT data (e.g., 1.3 million samples for HuatuoVision and 2 million for MedDR), our Base+RL model is built on a 7B base model using only 10K unannotated QA pairs.
For instance, our Base+RL model achieves an MMMU score of 57.33\%, significantly outperforming HuatuoVision’s 50.30\% on the same benchmark. 
In addition, on GMAI-MMbench, Base+RL obtains 43.14\% on the validation set and 43.84\% on the test set, surpassing MedDR’s scores of 41.95\% and 43.69\%, respectively. 
These results highlight that SFT not only enhances reasoning capabilities but does so efficiently, requiring far less computational and data-intensive resources compared to existing state-of-the-art models, which could be of significant benefit in resource-constrained scenarios such as medical applications.

\paragraph{Fine-grained Analysis of RLT.}

\input{src/tables/gmai_mmbench}

\begin{table*}[t]
  \centering
  \subfloat[
      \textbf{Inference strategy}. 
      \label{tab:scaling_factor}
  ]{
  \renewcommand{\arraystretch}{1.3}
      \begin{minipage}[t][3cm][c]{0.40\linewidth}{
          \begin{center}
              \resizebox{1\linewidth}{!}{
        \begin{tabular}{ll|cc}
            \hline
            Model & Type & MMMU(H\&M) & GMAI-MM(\textit{val}) \\
            \hline
            \multirow{2}{*}{Base(7B)}  & directly & 55.33 & 37.47 \\
             & cot & 55.33 & 40.02 \\
            \multirow{2}{*}{Base(7B)+RLT} & directly & 54.67 & 45.07 \\
             & cot & 57.73 & 43.14 \\ \hline
        \end{tabular}
    }
          \end{center}}
      \end{minipage}
  }
  \subfloat[
      \textbf{Model size}. 
      \label{tab:noise_schedule}
  ]{
  \renewcommand{\arraystretch}{1.3}
      \begin{minipage}[t][3cm][c]{0.335\linewidth}{
          \begin{center}
              \resizebox{1\linewidth}{!}{
              \tablestyle{1.5pt}{1.0}
              \begin{tabular}{l|cc}
              \hline
              \footnotesize{Type} & \footnotesize{MMMU(H\&M)} & \footnotesize{GMAI-MM(\textit{val})} \\
              \hline
              \footnotesize{Base(3B)} & \footnotesize{53.33} & \footnotesize{39.05} \\
              \footnotesize{Base(3B)+RLT} & \footnotesize{51.00} & \footnotesize{40.45} \\ 
              \footnotesize{Base(7B)} & \footnotesize{55.33} & \footnotesize{40.02} \\
              \footnotesize{Base(7B)+RLT} & \footnotesize{57.73} & \footnotesize{43.14} \\ \hline
              \end{tabular}
    }
          \end{center}}
      \end{minipage}
  }
  \subfloat[
      \textbf{Training steps (RLT)}.
      \label{tab:efficiency}
  ]{
      \begin{minipage}[t][3cm][c]{0.22\linewidth}{
          \begin{center}
              \resizebox{1\linewidth}{!}{
              \tablestyle{1pt}{1.0}
              \begin{tabular}{l|ccc}
              \hline
              \footnotesize{Step} & \footnotesize{MMMU(H\&M)} &\footnotesize{GMAI-MM(\textit{val})}   \\
              \hline
              \footnotesize{0} & \footnotesize{55.33}  & \footnotesize{40.02} \\
              \footnotesize{10} & \footnotesize{56.00} & \footnotesize{41.08} \\
              \footnotesize{20} & \footnotesize{54.00} &  \footnotesize{42.29} \\
              \footnotesize{100} & 
              \footnotesize{56.60} & \footnotesize{42.97} \\
              \footnotesize{165} & \footnotesize{57.73} & \footnotesize{43.14} \\ \hline
              \end{tabular}
              }
          \end{center}}
      \end{minipage}
  }
  \hspace{0.5em}
  \caption{Ablation experiments on the MMMU H\&M and the GMAI-MMBench \textit{val} set.}
  \label{tab:ablations}
\end{table*}

We further conduct fine-grained analysis on the GMAI-MMBench benchmark, which is a comprehensive medical multimodal benchmark designed to evaluate models on a range of clinical VQA tasks, as detailed in \cref{tab:gmai}. 
The results reveal that RLT significantly improves performance on more generalizable tasks, such as Disease Diagnosis (\textbf{DD}), Attribute Recognition (\textbf{AR}), and Organ Recognition (\textbf{OR-A/HN/P/T}). 
These tasks require nuanced analysis and reasoning, which RLT helps to improve.
For instance, RLT improves Disease Diagnosis performance from 47.42\% (Base+SFT) to 50.64\% (Base+RL). 
In a more extreme example, for counting (\textbf{C}) tasks, SFT leads to a 7.52\% drop in accuracy due to memorizing input-output mappings, which hinders the understanding of the underlying principles. 
However, RL boosts performance by 17\%, highlighting its advantage in tasks that require high-level reasoning and a deeper understanding of the image content, rather than simple memorization.
However, tasks like Surgical Instrument Recognition (\textbf{SIR}) show no significant performance difference between RLT and SFT, indicating that RL does not provide substantial benefits for more straightforward recognition tasks. 
This is likely because, unlike organs or attributes with varied representations, surgical instruments have a more uniform appearance. 
This suggests that for simpler recognition tasks, fine-tuning is sufficient to achieve competitive results, and RLT’s impact is less pronounced.

\begin{figure*}[t]
    \centering
    \includegraphics[width=\linewidth]{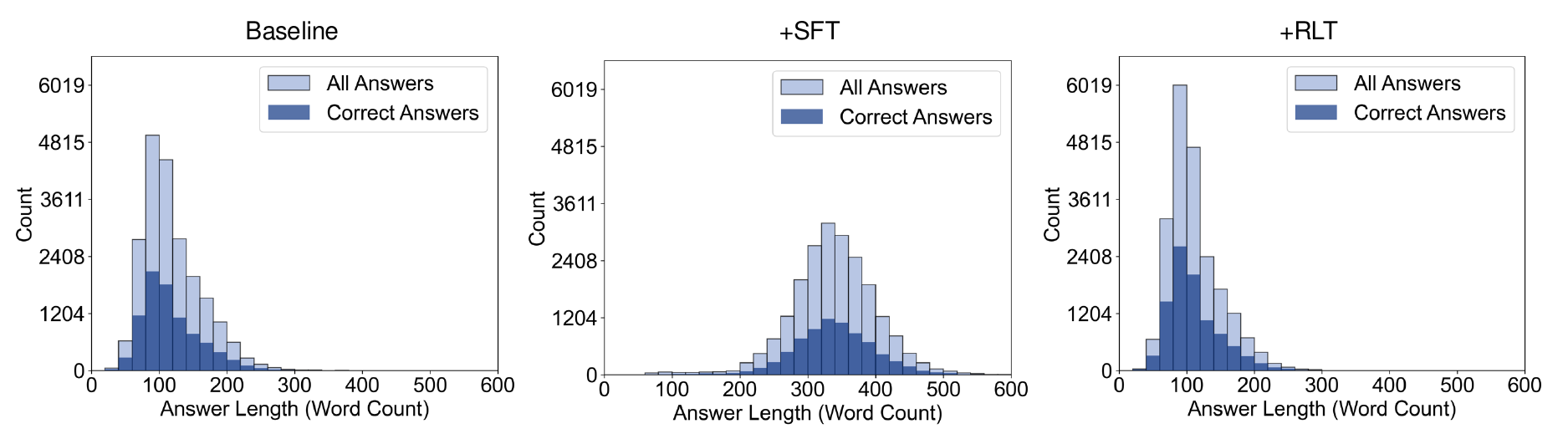}
    \caption{\textbf{Distribution of generated answer lengths (in word count) for the Baseline, +SFT, and +RLT models}. Each histogram displays the total number of answers (light bars) and correct answers (dark bars).}
    \label{fig:answer_length_distribution}
\end{figure*}

\paragraph{No ``Aha Moment" Yet.}
In the context of language models, ``aha moments" refer to self-validation behaviors where the model exhibits a sudden moment of clarity during its reasoning process. 
This phenomenon is often observed in complex reasoning tasks such as mathematical problem-solving or coding, where the model's chain-of-thought reveals explicit self-corrections and insights that lead to the final answer, and is accompanied by an increase in the length of the output answer tokens.
However, in our experiments with medical visual question-answering tasks, we did not observe such ``aha moments." 
The generated responses tend to have noticeably shorter reasoning chains (as shown in \cref{fig:answer_length_distribution}). This difference may stem from the inherently lower or less explicit reasoning demands of medical VQA tasks. 
Unlike math or coding, where the correct solution path is often obscured until the model re-evaluates its reasoning steps, medical VQA tasks typically require more direct associations between visual cues and clinical concepts. 
Consequently, the reasoning process in these tasks does not manifest the same clear-cut moments of self-validation, indicating a requirement for further exploration.

\subsection{More results}
\paragraph{} To provide a more detailed evaluation of the model's performance, we conducted an ablation study focusing on various key aspects, where the results are summarized in \cref{tab:ablations}.
\paragraph{Inference Strategy.} 
We compare two inference strategies: direct answer generation versus Chain of Thought (CoT)$-$based reasoning followed by answer generation. The results, summarized in \cref{tab:ablations} (a), show that while CoT reasoning can enhance the model's performance in some cases, it does not always lead to improvements. In particular, we observe that CoT reasoning slightly improves performance on GMAI-MMBench (\textit{val}), increasing the score from 37.47\% to 40.02\%, but it does not consistently outperform direct answer generation across all settings.
After applying RLT, both inference strategies show significant improvements. Direct answer generation achieves scores of 54.67\% on MMMU and 45.07\% on GMAI-MMBench, whereas CoT reasoning leads to 57.73\% on MMMU but a slight drop to 43.14\% on GMAI-MMBench. Upon further analysis, we found that in some cases, CoT reasoning negatively impacted performance due to excessively long or repetitive outputs, which prevented the model from generating the correct final answer. These findings highlight that while structured reasoning can be beneficial, it must be carefully controlled to avoid unnecessary verbosity or redundancy that could degrade performance.

\paragraph{Model Size.}
To evaluate the impact of model size, we compare the performance of two model variants: Qwen2.5-VL-3B and Qwen2.5-VL-7B, both before and after RLT.
After RLT, the 7B model shows a notable improvement, increasing by 2.40\% on MMMU and 3.12\% on GMAI-MMBench, while the 3B model improves by 1.40 points on GMAI-MMBench but decreases by 2.33\% on MMMU. The results suggest that the 7B model benefits more from reinforcement learning (RL) than the 3B model. This may be because the 3B model has a weaker foundational capability in medical tasks, making it harder to effectively leverage RLT feedback for significant performance improvements. However, the 7B model with stronger performance, can better utilize reinforcement signals to refine its reasoning and decision-making abilities. This disparity highlights the importance of model scaling when applying RL.

\paragraph{Training Steps.}
To evaluate the effect of the RLT training steps, we assess the performance of the 7B model at various intervals: 0, 10, 20, 100, and 165 steps. The results, summarized in \cref{tab:ablations}~(c), demonstrate a generally stable improvement in performance as the number of training steps increases.
After 10 RLT steps on the base model, the scores increase by 0.67\% on MMMU and 1.06\% on GMAI-MMBench, showing that even a small amount of RL feedback can enhance performance. Further evaluations over multiple steps reveal that performance on the GMAI-MMBench dataset consistently improves, showing relatively stable growth. However, on the MMMU dataset, the performance exhibits fluctuations, which may be attributed to the smaller size of the benchmark, making it more prone to variability.
These results highlight the stability of our RLT method, as longer training enables the model to consistently refine its decision-making and reasoning capabilities.

%% file: src/tables/overall.tex
\renewcommand{\arraystretch}{1.2}
\begin{table*}[t]
    \centering
    \resizebox{\textwidth}{!}{
    \begin{tabular}{l|>{\centering}p{2.6cm}>{\centering}p{2.6cm}cccc}
    \hline
    Method       & MMMU  & MMMU-pro & GMAI-MMbench(\textit{val}) & GMAI-MMbench(\textit{test}) & MedXpertQA-MM & OmniMedVQA \\ \hline
    Med-Flamingo (7B)~\cite{moor2023med} & 28.40 & $\_$          & 12.74             & 11.64              & $\_$               & 23.82      \\
    RadFM (13B)~\cite{wu2023towards}        & 27.90 & $\_$          & 22.95             & 22.93              & $\_$               & 23.48      \\
    LLaVA-Med (7B)~\cite{li2024llava}    & 38.60 & $\_$          & 20.54             & 19.60              & $\_$               & 27.82      \\
    HuatuoVision (34B)~\cite{chen2024huatuogpt} & 50.30 & $\_$          & $\_$                   & $\_$                    & $\_$               & \textbf{73.23}      \\
    MedDR (40B)~\cite{he2024meddr}        &  $\_$      & $\_$          & 41.95             & 43.69              & $\_$               & 68.27      \\ \hline
    Base (Qwen2.5-vl-7B) & 55.33 & 28.47 & 40.02      & 40.59              & 20.30          & 58.41      \\
    Base+SFT     & 56.00 & 32.99 & 39.65             & 36.12              & 23.55          & 66.34      \\
    \rowcolor{gray!20} 
    Base+RLT      & \textbf{57.33} & \textbf{34.03} & \textbf{43.14}             & \textbf{43.84}              & \textbf{23.80}          & 61.01      \\
    $\Delta$ & \textcolor{blue}{\textbf{+2.00}} & \textcolor{blue}{\textbf{+5.56}} & \textcolor{blue}{\textbf{+3.12}} & \textcolor{blue}{\textbf{+3.25}} & \textcolor{blue}{\textbf{+3.50}} & \textcolor{blue}{\textbf{+2.60}} \\ \hline
    \end{tabular}}
    \caption{\textbf{Comparison of different tuning strategies across multiple medical multimodal benchmarks, along with state-of-the-art methods}. The bottom row ($\Delta$) indicates the performance gain over the base model. The best results are bolded.}
    \label{tab:overall_comparsion}
\end{table*}

%% file: src/tables/gmai_mmbench.tex
\begin{table*}[t]
    \centering
    \resizebox{\linewidth}{!}{
    \setlength{\tabcolsep}{3.5pt}
    \begin{tabular}{l|cc|cccccccccccccccccc}
    \hline
    \multirow{2}{*}{\textbf{Model Name}}  
    & \multirow{2}{*}{\makecell{\textbf{Overall}\\\textit{(val)}}}
    & \multirow{2}{*}{\makecell{\textbf{Overall}\\\textit{(test)}}}
    & \multirow{2}{*}{\textbf{AR}} & \multirow{2}{*}{\textbf{BVR}} & \multirow{2}{*}{\textbf{B}} 
    & \multirow{2}{*}{\textbf{CR}} & \multirow{2}{*}{\textbf{C}}  & \multirow{2}{*}{\textbf{DD}} 
    & \multirow{2}{*}{\textbf{IQG}}& \multirow{2}{*}{\textbf{MR}} & \multirow{2}{*}{\textbf{M}}  
    & \multirow{2}{*}{\textbf{NT}} & \multirow{2}{*}{\textbf{OR-A}}& \multirow{2}{*}{\textbf{OR-HN}} 
    & \multirow{2}{*}{\textbf{OR-P}}& \multirow{2}{*}{\textbf{OR-T}}& \multirow{2}{*}{\textbf{SG}}  
    & \multirow{2}{*}{\textbf{SAR}}& \multirow{2}{*}{\textbf{SIR}}& \multirow{2}{*}{\textbf{SWR}}\\
     & & | \\
    \midrule
    Random   & 25.70 & 25.94 & 38.20 & 22.73 & 22.92 & 22.72 & 24.06 & 26.66 & 27.13 
             & 27.00 & 20.00 & 24.75 & 21.37 & 22.93 & 22.33 & 21.18 & 32.43 & 24.23 
             & 21.39 & 23.71 \\
    \midrule
    Claude3-Opus~\citep{anthropic2024claude}
             & 32.37 & 32.44 & 1.61  & 39.51 & 34.31 & 31.66 & 12.63 & 39.26 & 28.74
             & 30.86 & 22.40 & 37.37 & 25.79 & 41.07 & 29.33 & 33.18 & 31.31 & 21.35
             & 23.87 & 4.00 \\
    Qwen-VL-Max~\citep{bai2023qwen}
             & 41.34 & 42.16 & 32.68 & 44.58 & 31.38 & 40.79 & 10.68 & 50.53 & 32.79
             & 44.36 & 29.20 & 51.52 & 41.37 & 58.00 & 30.67 & 41.65 & 26.95 & 25.00
             & 24.64 & 39.14 \\
    GPT-4V~\citep{achiam2023gpt}
             & 42.50 & 44.08 & 29.92 & 48.95 & 44.00 & 37.39 & 12.93 & 52.88 & 32.79
             & 44.21 & 32.80 & 63.64 & 39.89 & 54.13 & 37.00 & 50.59
             & 27.55 & 23.08 & 25.75 & 37.43 \\
    Gemini 1.0~\citep{team2023gemini}
             & 44.38 & 44.93 & 42.12 & 45.10 & 46.46 & 37.57 & 20.45
             & 53.29 & 35.22 & 36.94 & 25.20 & 51.01 & 34.74 & 59.60 & 34.00 & 50.00
             & 36.64 & 23.65 & 23.87 & 35.43 \\
    Gemini 1.5~\citep{reid2024gemini}
             & 47.42 & 48.36 & 43.50 & 56.12
             & 51.23 & 47.58 & 2.26  & 55.33
             & 38.87 & 48.07 & 30.00 & 76.26
             & 51.05 & 75.87 & 46.33 & 62.24 & 20.57 & 27.69
             & 30.54 & 40.57 \\
    GPT-4o~\citep{achiam2023gpt}
             & 53.53 & 53.96 & 38.32 & 61.01 & 57.08
             & 49.02 & 46.62 & 61.45 
             & 46.56 & 56.38 & 34.00 & 75.25 & 53.79 & 69.47 & 48.67
             & 65.88 & 33.93 & 22.88 & 29.51 & 39.43 \\
    \midrule
    Med-Flamingo~\citep{moor2023med}
             & 12.74 & 11.64 & 6.67  & 10.14 & 9.23  & 11.27 & 6.62  & 13.43 & 12.15
             & 6.38  & 8.00  & 18.18 & 9.26  & 18.27 & 11.00 & 11.53 & 12.16 & 5.19
             & 8.47  & 11.43 \\
    LLaVA-Med~\citep{li2024llava}
             & 20.54 & 19.60 & 24.51 & 17.83 & 17.08 & 19.86 & 15.04 & 19.81 & 20.24
             & 21.51 & 13.20 & 15.15 & 20.42 & 23.73 & 17.67 & 19.65 & 21.70 & 19.81
             & 14.11 & 20.86 \\
    Qilin-Med-VL-Chat~\citep{liu2023qilin}
             & 22.34 & 22.06 & 29.57 & 19.41 & 16.46 & 23.79 & 15.79 & 24.19 & 21.86
             & 16.62 & 7.20  & 13.64 & 24.00 & 14.67 & 12.67 & 15.53 & 26.13 & 24.42
             & 17.37 & 25.71 \\
    RadFM~\citep{wu2023towards}
             & 22.95 & 22.93 & 27.16 & 20.63 & 13.23 & 19.14 & 20.45 & 24.51 & 23.48
             & 22.85 & 15.60 & 16.16 & 14.32 & 24.93 & 17.33 & 21.53 & 29.73 & 17.12
             & 19.59 & 31.14 \\
    MedDr~\citep{he2024meddr}
             & 41.95 & 43.69 & 41.20 & 50.70 & 37.85 & 29.87 & 28.27 & 52.53 & 36.03
             & 31.45 & 29.60 & 47.47 & 33.37 & 51.33 & 32.67 & 44.47 & 35.14 & 25.19
             & 25.58 & 32.29 \\
    \midrule
    \textbf{Base(Qwen2.5-VL-7B)}
             & 40.02 & 40.55 & 44.89 &  46.21 &  34.37 &  37.79 &  9.47 &  47.42 &  35.60 &  36.35 &  23.20 &  44.00 &  38.35 &  45.55 &  35.20 &  37.53 &  32.85 &  24.23 &  28.38 &  37.71 \\
    \textbf{Base+SFT}
             & 39.65 & 36.12 & 41.33 &  38.49 &  31.84 &  39.57 &  1.95 &  41.60 &  31.20 &  22.11 &  22.80 &  47.50 &  39.21 &  41.42 &  31.20 &  33.76 &  31.28 &  24.62 &  25.91 &  30.29 \\
    \rowcolor{gray!20}
    \textbf{Base+RLT}
             & 43.14 & 43.84 &  51.41 &  49.00 &  35.63 &  41.35 &  26.47 &  50.64 &  37.20 &  37.54 &  25.60 &  49.00 &  39.55 &  49.94 &  40.00 &  42.24 &  33.81 &  25.58 &  28.29 &  40.57  \\
    $\Delta$ &  \textcolor{blue}{\textbf{+3.12}} & \textcolor{blue}{\textbf{+3.29}} & \textcolor{blue}{\textbf{+6.52}} & \textcolor{blue}{\textbf{+2.79}} & \textcolor{blue}{\textbf{+1.26}} & \textcolor{blue}{\textbf{+3.56}} & \textcolor{blue}{\textbf{+17.00}} & \textcolor{blue}{\textbf{+3.22}} & \textcolor{blue}{\textbf{+1.60}} & \textcolor{blue}{\textbf{+1.19}} & \textcolor{blue}{\textbf{+2.40}} & \textcolor{blue}{\textbf{+5.00}} & \textcolor{blue}{\textbf{+1.20}} & \textcolor{blue}{\textbf{+4.39}} & \textcolor{blue}{\textbf{+4.80}} & \textcolor{blue}{\textbf{+4.71}} & \textcolor{blue}{\textbf{+0.96}} & \textcolor{blue}{\textbf{+1.35}} & -0.09 & \textcolor{blue}{\textbf{+2.86}}
    \\ 
    \hline
    
    
    \end{tabular}
    }
    \caption{\textbf{Results on the \textit{val} and \textit{test} sets of GMAI-MMBench for clinical VQA tasks}.The sub-items behind are evaluated on the test set.  
    The full names of the evaluated tasks can be found in Table 5 in literature~\citep{chen2024gmai}. 
    }
    \label{tab:gmai}
    \end{table*}

%% file: src/secs/5_conclusion.tex
\section{Conclusion}

In this paper, we introduce GMAI-VL-R1, an innovative multimodal medical reasoning model that leverages reinforcement learning tuning to enhance its reasoning and reflective capabilities. Compared to existing models, GMAI-VL-R1 optimizes its decision-making process through long-chain reasoning and integrates a reflection mechanism to fine-tune its reasoning results, significantly improving diagnostic accuracy and clinical decision support. To address the reasoning challenges in complex medical decision-making, we propose a multi-agent reasoning data synthesis approach, utilizing rejection sampling to generate preliminary reasoning data and employing another agent to reflect and adjust the generated data, thus improving the model's reasoning quality and generalization ability.

Experimental results demonstrate that GMAI-VL-R1 outperforms current state-of-the-art multimodal medical models across several benchmark tasks, particularly in complex reasoning tasks such as medical image diagnosis and visual question answering. The success of GMAI-VL-R1 highlights the critical role of reinforcement learning in multimodal medical reasoning, enabling it to better tackle the challenges of complex clinical decision-making.

%% file: src/secs/8_appendix.tex
\section{Appendix Title}

\begin{table}[htbp]
    \centering
    \caption{Training settings of GMAI-VL-R1's Stage I (SFT) and Stage II (RL).}
    \label{tab:setting}
    \resizebox{1.0\linewidth}{!}{
    \begin{tabular}{l|cc}
    \hline
    \textbf{Settings}               & \textbf{SFT} & \textbf{RL} \\ \hline
    freeze LLM                      & False             & False            \\ 
    freeze MLP                      & False            & False            \\ 
    freeze Vision Encoder           & False             & False            \\ 
    learning rate                   & 1e-4             & 1e-5             \\ 
    optimizer                       & AdamW            & AdamW            \\ 
    optimizer hyper-parameters      & $\beta_1=0.9, \beta_2=0.999$ & $\beta_1=0.9, \beta_2=0.999$  \\ 
    total batch size                & 8x4x8          & 8x1x8           \\ 
    drop rate                       & 0.0              & 0.0              \\ 
    numerical precision             & DeepSpeed bf16   & DeepSpeed bf16   \\ 
    GPUs for training               & 8xA100 (80G)    & 8xA100 (80G)    \\ 
    \hline
    \end{tabular}}
\end{table}

\subsection{benchmarks}
\label{sec:benchmark}
Below is a brief overview of the benchmarks used in our experiments:

\begin{itemize}
    \item \textbf{MMMU Health \& Medicine track}: 
    The Health \& Medicine track of the MMMU~\cite{yue2024mmmu} benchmark spans a wide range of medical fields, derived from university exams, quizzes, and textbooks. It evaluates the model's reasoning ability in complex medical scenarios and the specialized knowledge in health and medicine.
    \item \textbf{MMMU-Pro Health \& Medicine track}:MMMU-Pro~\cite{yue2024mmmupro} is an enhanced version of the MMMU~\cite{yue2024mmmu} benchmark, designed to test multimodal models' reasoning by filtering text-only questions, expanding answer options, and adding vision-only input. It better mimics real-world scenarios, requiring integration of visual and textual information, and reveals current models' limitations.
    
    \item \textbf{OmniMedVQA}:  
    OmniMedVQA~\cite{hu2024omnimedvqa} provides a rich dataset of paired medical images and text, designed to evaluate the model's ability to recognize and understand fundamental medical imaging concepts, focusing on cross-modal reasoning and information integration.
    
    \item \textbf{GMAI-MMBench}:  
    GMAI-MMBench~\cite{chen2024gmai} focuses on assessing the model's ability to identify fine-grained objects in complex clinical scenarios, challenging its capacity to handle long-context tasks and accurately recognize and reason over detailed medical features. In addition, it consists of both validation and test sections.

    \item \textbf{MedXpertQA-MM}:
    MedXpertQA~\cite{medxpert} is a comprehensive and challenging benchmark designed to evaluate expert-level medical knowledge and advanced reasoning, introducing expert-level exam questions with diverse clinical information, including patient records and examination results, distinguishing it from traditional medical multimodal benchmarks.

\end{itemize}

\subsection{system promot}

\begin{table}[h]
\centering
\caption{System prompt}
\begin{tabular}{{p{\linewidth}}} 
\hline
A conversation between User and Assistant. The user asks a question, and the Assistant solves it. The assistant first thinks about the reasoning process in the mind and then provides the user a concise final answer in a short word. The reasoning process and answer are enclosed within \textless think\textgreater \text{ reasoning process here } \textless /think\textgreater and \textless answer\textgreater \text{ answer here } \textless /answer\textgreater tags, respectively, i.e., \textless think\textgreater reasoning process here \textless /think\textgreater\textless answer\textgreater answer here \textless /answer\textgreater" \\
\hline
\end{tabular}
\end{table}


\begin{table*}[h!]
\centering
\caption{Prompt Template for Medical Imaging Multi-Choice Question Construction}
\begin{tabular}{{p{\linewidth}}}
\hline
\textbf{QUESTION:} Based on the given medical image and the answer set, propose a question whose answer can be inferred from the image. The correct answer to the question must come from one of the elements in the answer set; no new or extended answers may be added. \\
\textbf{CHOICE:} Label each option sequentially with letters (A, B, C, D, …). Each option’s content must strictly match one element in the answer set, without repetition or new additions. If the answer set has more than 4 items, randomly select 4 of them; if it has fewer than 4, only create as many options as exist. Ensure that the correct answer is among these options.\\
\textbf{ANALYSIS:} First, provide an overview of your solution strategy: how you combine image and clinical knowledge to make a judgment. Then, explain the reasoning process step by step: at each step, detail what information you derive from the image and from medical knowledge, and what conclusions you draw. Finally, analyze each option in detail, explaining whether or not it could be the correct answer, and arrive at the final conclusion. \\
\textbf{ANSWER:} Provide only the letter corresponding to the correct option (e.g., “A” or “B”), without repeating the text of that option. \\
\textbf{Sample Output} (json format):\\
\texttt{\raggedright
\{ \\
    "QUESTION": "Enter the question here, which should be inferable from the image", \\
    "CHOICE": \{ \\
        "A": "Content of option A", \\
        "B": "Content of option B", \\
        "C": "Content of option C", \\
        "D": "Content of option D" \\
    \}, \\
    "ANALYSIS": "Enter the detailed analysis here", \\
    "ANSWER": "Letter corresponding to the correct option" \\
\}
} \\
\hline
\end{tabular}
\end{table*}


\begin{table*}[h!]
\centering
\caption{Prompt Template for Medical Imaging Question Construction}
\begin{tabular}{p{\textwidth}}
\hline
As a professional medical imaging expert, your responsibility is to thoroughly explore medical images and provide accurate, precise answers based on your clinical expertise. You will be given an image and a set of possible answers, and your task is to construct a question and provide a reasoned answer. \\
The reasoning process should be enclosed within \textless think\textgreater tags, and the final answer should be enclosed within \textless answer\textgreater tags. Additionally, the question should be enclosed within \textless question\textgreater. \\
\textbf{Steps:} \\
1. \textbf{Question Construction:} \\
Based on the provided image and the possible answers, construct a question that can be answered based on the image. The question must be clear and directly relevant to the information in the image. \\

2. \textbf{Reasoning Process:} \\
- Analyze the image step by step without referencing the possible answers directly. \\
- Provide the step-by-step reasoning process, explaining the rationale and supporting each step. \\
- Consider the clinical knowledge required and how the image provides insight into the condition being examined. \\
- Clearly explain the steps of the reasoning process that lead to the final conclusion. \\

3. \textbf{Final Answer:} \\
- After reasoning, provide the final answer derived from the image analysis. This must be one of the options in the answer set. The answer should be concise and logically follow from the reasoning process. \\

\textbf{Input Information:} \\
- Imaging Modality: \{modality\} \\
- Background: \{knowledge\} \\
- Answer Set: \{answer\_set\} \\

\textbf{Output Format:} \\ 
\textless question\textgreater Your question here, based on the image and answer set \textless /question\textgreater \\
\textless think \textgreater Your detailed reasoning process, step by step \textless /think\textgreater \\
\textless answer \textgreater Your final answer, matching one of the options in the answer set \textless /answer\textgreater \\
\\
\hline
\end{tabular}
\end{table*}